\documentclass[lettersize,journal]{IEEEtran}
\usepackage{amsmath,amsfonts}
\usepackage{algorithmic}
\usepackage{algorithm}
\usepackage{array}
\usepackage{textcomp}
\usepackage{stfloats}
\usepackage{subfloat}
\usepackage{url}
\usepackage{verbatim}
\usepackage{subfloat}
\usepackage{cite}
\usepackage{gensymb}
\usepackage{amssymb}
\usepackage{color}
\usepackage{makecell} 
\usepackage{graphicx}  
\usepackage[mathscr]{eucal}
\usepackage[caption=false,font=scriptsize,labelfont=rm,textfont=rm]{subfig}
\usepackage{booktabs}
\usepackage{float}   
\usepackage{ragged2e} 
\usepackage{bbm}
\usepackage{bm}
\usepackage{xcolor}
\usepackage{hyperref}
\usepackage{multirow}
\usepackage{pifont}  
\newcommand{\cmark}{\ding{51}}  
\newcommand{\xmark}{\ding{55}}  

\hypersetup{hypertex=true,
	colorlinks=true,
	linkcolor=blue,
	anchorcolor=blue,
	citecolor=blue}
\usepackage{orcidlink}
\usepackage{tikz}
\begin{document}

\title{PromptMID: Modal Invariant Descriptors Based on Diffusion and Vision Foundation Models for Optical-SAR Image Matching} 

\author{Han Nie\hspace{-1.0mm}$^{~\orcidlink{0000-0001-9229-6109}}$, 
Bin Luo\hspace{-1.0mm}$^{~\orcidlink{0000-0002-3040-3500}}$,~\IEEEmembership{Senior Member,~IEEE},
 Jun Liu\hspace{-1.0mm}$^{~\orcidlink{0000-0002-8943-079X}}$,
Zhitao Fu\hspace{-1.0mm}$^{~\orcidlink{0000-0002-1212-7186}}$,
  Huan Zhou\hspace{-1.0mm}$^{~\orcidlink{0009-0004-8146-5076}}$,
  Shuo Zhang\hspace{-1.0mm}$^{~\orcidlink{0009-0004-5002-4577}}$,
 Weixing Liu\hspace{-1.0mm}$^{~\orcidlink{0000-0002-0681-5257}}$

\thanks{
	This work was supported by the National Natural Science Foundation of China under Grant 41961053 and Yunnan Fundamental Research Projects under Grant 202301AT070463.	
	\IEEEcompsocthanksitem Han Nie, Jun Liu, Shuo Zhang, Weixing Liu and Bin Luo are with the State Key Laboratory of Information Engineering in Surveying, Mapping and Remote Sensing, Wuhan University, Wuhan 430079, China (e-mail:liujunand@whu.edu.cn).
	\IEEEcompsocthanksitem Huan Zhou is with the School of Remote Sensing and Information Engineering, Wuhan University, Wuhan 430079, China.
	\IEEEcompsocthanksitem Zhitao Fu is with the Faculty of Land Resources Engineering, Kunming University of Science and Technology, Kunming 650031, China.
	\IEEEcompsocthanksitem Corresponding author: Jun Liu; Zhitao Fu.}
}
\maketitle
\begin{abstract} 
The ideal goal of image matching is to achieve stable and efficient performance in unseen domains. However, many existing learning-based optical-SAR image matching methods, despite their effectiveness in specific scenarios, exhibit limited generalization and struggle to adapt to practical applications. Repeatedly training or fine-tuning matching models to address domain differences is not only not elegant enough but also introduces additional computational overhead and data production costs. In recent years, general foundation models have shown great potential for enhancing generalization. However, the disparity in visual domains between natural and remote sensing images poses challenges for their direct application. Therefore, effectively leveraging foundation models to improve the generalization of optical-SAR image matching remains challenge. To address the above challenges, we propose PromptMID, a novel approach that constructs modality-invariant descriptors using text prompts based on land use classification as priors information for optical and SAR image matching. PromptMID extracts multi-scale modality-invariant features by leveraging pre-trained diffusion models and visual foundation models (VFMs), while specially designed feature aggregation modules effectively fuse features across different granularities. Extensive experiments on optical-SAR image datasets from four diverse regions demonstrate that PromptMID outperforms state-of-the-art matching methods, achieving superior results in both seen and unseen domains and exhibiting strong cross-domain generalization capabilities. The source code will be made publicly available \url{https://github.com/HanNieWHU/PromptMID}.
\end{abstract}


\section{Introduction}
\IEEEPARstart{W}{ith} the rapid development of remote sensing technology, a significant volume of multi-sensor, multi-resolution, and multi-temporal ground observation data has been acquired. Owing to the complementary characteristics of different sensor data, the co-processing of multi-sensor data demonstrates significant potential, particularly in the fields of change detection~\cite{9057421}, image fusion~\cite{TANG2023101870}, and land use classification~\cite{LI2023272}. Consequently, the joint processing of multi-modal images has garnered extensive attention and been studied extensively. Extracting advantageous and complementary information from multi-modal images to enhance the performance of downstream tasks has become a critical focus~\cite{li2024sm3det,10145843}. However, matching between multi-modal images remains one of the core challenges. In particular, geometric deformation and nonlinear radiation differences (NRDs) arising from differences in the imaging mechanisms of optical and synthetic aperture radar (SAR) images significantly hinder matching performance.

\begin{figure}[!t]
	\centering
	\includegraphics[width=\linewidth]{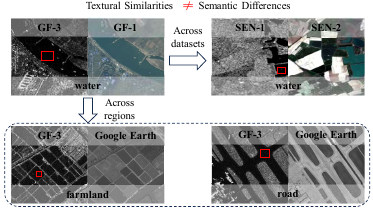}
	\caption{In optical and SAR image matching across datasets and regions, the textural and semantic ambiguity of SAR images hinders the extraction of modality-invariant features, resulting in decreased accuracy when matching in the unseen domain. The upper text in the image represents the image source, while the lower text provides semantic information corresponding to the red box.}
	\label{fig.question}
\end{figure}

To address the above challenges, researchers have introduced two main traditional hand-crafted approaches: area-based and feature-based methods. area-based methods~\cite{ye2017robust,ye2019fast,ye2022robust} use similarity metrics such as mean absolute difference (MAD) or mutual information (MI)~\cite{suri2009mutual} to estimate correspondences between images, but they struggle with complex matching tasks involving rotation or scaling transformations. In contrast, feature-based methods~\cite{lowe2004distinctive} construct matching pipelines to generate invariant descriptors after keypoint detection, thereby enhancing robustness to rotation and scaling transformations. However, homologous image matching approaches struggle to address the NRDs issue in multimodal images. To mitigate significant modal differences, some methods extract structural or phase information to construct descriptors~\cite{8935498,li2022lnift,FAN2024102252}. However, these methods rely on fixed, hand-crafted feature extractors, limiting their ability to capture shared information and resulting in suboptimal matching performance.

\begin{figure*}[!t]
	\centering
	\includegraphics[width=\linewidth]{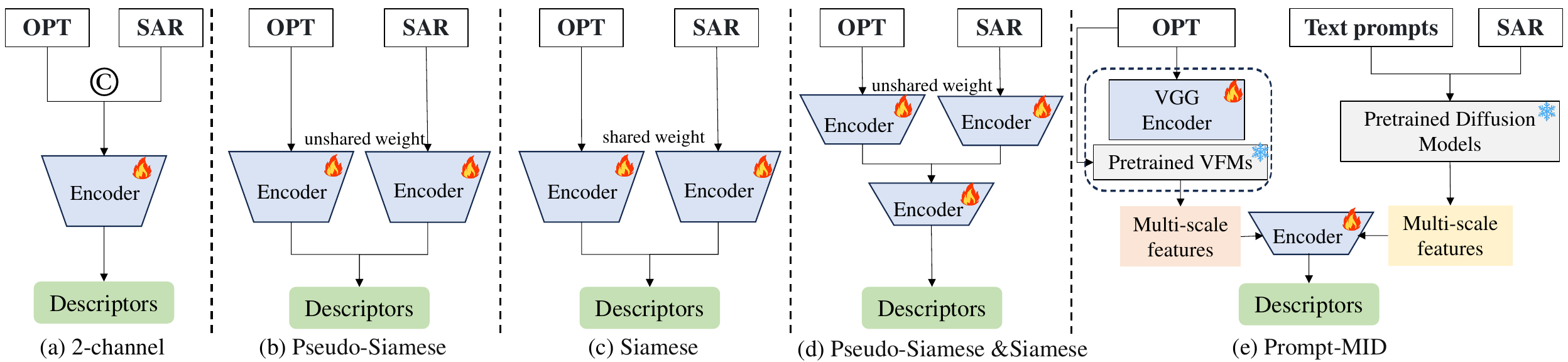}
	\caption{From left to right, various methods for constructing descriptors in optical and SAR image matching are presented. (a) Optical and SAR images are concatenated and then passed through an encoder to extract descriptors. (b) Descriptors are extracted by a Pseudo-Siamese network with unshared weights. (c) Descriptors are extracted by a Siamese network with shared weights. (d) Non-aligned features are extracted by Pseudo-Siamese networks with unshared weights, and aligned features are extracted by Siamese networks with shared weights. (e) The proposed Prompt-MID uses pre-trained vision foundation models and diffusion models as feature extractors, incorporates text prompts to extract modality-invariant features, and constructs robust descriptors. The snowflake icon in the figure indicates parameter freezing, and the fire symbol indicates parameter training.}
	\label{fig.ggg}
\end{figure*}

In recent years, the rapid advancement of deep learning has enabled learning-based methods~\cite{mishchuk2017working,tian2020hynet,sun2021loftr} to demonstrate greater potential compared to traditional hand-designed methods. Learning-based methods~\cite{9999700,tuzcuouglu2024xoftr,han2024contrastive,ye2024robust,quan2022self,quan2023efficient} are typically trained on rich unimodal datasets, where descriptors are constructed by extracting deep features from optical and SAR images. While these methods achieve superior performance on training datasets, as illustrated in Fig.~\ref{fig.ggg} (a)-(d), they often exhibit limited generalization capabilities when applied to unseen domains. Some methods attempt to train independent models on different regions or datasets, but these approaches face the following limitations: i) manual annotation of matched correspondences is both time-consuming and complex, making it challenging to obtain large-scale datasets with accurate ground truth; ii) The limited coverage of training data scenarios, combined with the absence of semantic guidance, leading to an under-representation of modality-invariant features, significantly degrades generalization performance across regions and datasets, as illustrated in Fig.~\ref{fig.question}. Currently, existing methods ~\cite{10064578,10662912,10282673} primarily rely on Generative Adversarial Networks (GANs) for modality unification. However, these methods are prone to local minima~\cite{diff_sar}, leading to instability in the transformation process for unseen domain images, which further reduces matching accuracy and generalization capability, ultimately constraining their practical applicability.

Recently, foundation models such as GPT-4~\cite{achiam2023gpt}, SAM~\cite{kirillov2023segany}, DINOv2~\cite{oquab2023dinov2}, and CLIP~\cite{radford2021learning} have achieved significant breakthroughs in natural imagery, with their strong generalization capabilities being a key factor enabling practical applications. Foundation models tailored for remotely sensed images remain in their infancy, msGFM~\cite{han2024bridging} effectively integrates data from four major sensor modalities to construct multisensor geospatial pretraining models. The primary objective of this paper is not to develop a generalised foundation models for optical and SAR images, but rather to explore the applicability of foundation models from natural imagery to optical and SAR image matching downstream tasks. Direct application of foundation models from the natural image domain to remote sensing images often results in degraded accuracy and significant limitations, primarily due to the inherent differences between the visual domains of natural and remote sensing images~\cite{ZERMATTEN2025621}. The main objective of this study is to transfer the foundation model from the natural image domain to the remote sensing image domain through representation learning and apply it to optical and SAR matching in order to learn modal invariant features and improve the matching accuracy. By leveraging the foundation model, this research aims to improve the generalization of the model to unseen domains and to promote the continuous progress and practical application of optical and SAR image matching.

To enhance the generalization ability of optical and SAR image matching through foundation models, we propose a novel method, called PromptMID, designed to constructing modality-invariant descriptors. Our approach is primarily based on the diffusion models~\cite{SD} and visual foundation models (VFMs)~\cite{oquab2023dinov2}, as illustrated in Fig.~\ref{fig.ggg} (e).
Additionally, text prompts are generated based on land use classification, with semantic information describing SAR images serving as priors information to guide the extraction of modality-invariant features. Two major challenges arise in this process: 1) How to extract robust and reliable modality-invariant features from optical and SAR images? 2) How to utilize these extracted modality-invariant features to construct discriminative descriptors? 

To address these challenges, we propose text prompts based on land use classification as priors information to guide the diffusion model in extracting diffusion features as modality-invariant representations. Specifically, we utilize the intermediate features of the diffusion decoder as multi-scale latent diffusion features in the mapping process from text and SAR images to optical images. For optical images, we extract coarse-grained features from frozen VFMs and fine-grained features from learnable VGG models, integrating both to construct multi-scale representations. However, the diffusion features of SAR images themselves contain noise, while there are significant differences in the field of view between remote sensing imagery and natural imagery, which, if directly used to construct descriptors, would weaken the discriminative and invariant properties of the features, thus significantly reducing the matching accuracy. To this end, we propose a multi-scale aware aggregation module (MSAA) that effectively integrates features across different scales and fuses information at various granularities through aware mechanism, thereby enhancing global representation capability and improving the matching accuracy. However, multi-scale aggregation may also introduce information redundancy. To mitigate this issue, we incorporate Convolutional Block Attention Module (CBAM)~\cite{cbam} that adaptively refines features in both the channel and spatial dimensions, effectively suppresses irrelevant feature interference while enhancing feature representation.

The contributions of this paper can be summarized as follows:
\begin{itemize}
	\item
    We present PromptMID, a method for constructing modality-invariant descriptors using text prompts based on land use classification as priors information for optical and SAR image matching. To the best of our knowledge, PromptMID is the first approach to incorporate land use classification information as a priori information for optical and SAR image matching.
		
\item
    PromptMID significantly improves the generalization of descriptors by combining pre-trained diffusion models and VFMs, demonstrating the potential of the foundation model in optical and SAR image matching. Meanwhile, we design multi-scale aware aggregation module (MSAA) to effectively fuse features with different granularities. The discriminative and invariant nature of the descriptors is further enhanced by the introduction of CBAM module to refine the features.
		
\item
    We conducted extensive experiments on four distinct optical and SAR datasets, cross dataset has different image size, imaging sensors, or geographic regions. The experimental results demonstrate that our method exhibits strong generalization capabilities across regions and datasets.
\end{itemize}

\section{Related Work}
\label{2}
\subsection{Traditional handcrafted methods}
In the last decades, traditional handcrafted methods have been widely developed and can be divided into two categories: area-based methods and feature-based methods.

{\bf{Area-based methods:}} The core of area-based alignment methods lies in designing an appropriate similarity metric and determining the optimal matching position through a template-matching strategy for image alignment optimization. Commonly used similarity metrics in area-based methods include normalized cross-correlation (NCC), sum of squared differences (SSD), and mutual information (MI). However, SSD is highly sensitive to noise, making it unsuitable for optical and SAR image matching. In contrast, NCC is robust to illumination changes and effective for translational shifts but is highly sensitive to rotational variations. Due to its robustness to nonlinear radiometric differences, MI is widely used for multimodal image matching~\cite{OFVERSTEDT2022196}; however, it is prone to local optima. To address the NRDs in multimodal images, methods such as histogram of oriented phase congruency (HOPC)~\cite{ye2017robust} and channel features of orientated gradients (CFOG)~\cite{ye2019fast} introduce more robust region descriptors. While these methods perform well in scenarios without rotational transformations, they are susceptible to geometric distortions, exhibit poor generalization, and are challenging to apply widely. Although these approaches demonstrate improved performance on specific datasets, they often struggle with complex geometric distortions, limiting their effectiveness in broader applications.

{\bf{Feature-based methods:}} 
Over the past few decades, numerous classical feature-based matching methods have been developed, typically comprising three main stages: keypoint detection, description, and matching. Methods such as scale-invariant feature transform (SIFT)~\cite{lowe2004distinctive} and speeded up robust features (SURF)~\cite{bay2008speeded} rely on gradient-based designs, making them challenging to apply to NRDs. To address this limitation, radiation-invariant feature transform (RIFT)~\cite{8935498} extracts maximum index map (MIM) maps based on phase consistency, offering improved feature robustness. Building on this, approaches such as locally normalized image feature transform (LNIFT)~\cite{li2022lnift} and scale and rotation invariant feature transform (SRIF)~\cite{li2023multimodal} have been introduced to enhance computational efficiency and improve descriptor robustness against rotation and scale transformations. Although these methods have achieved success on multimodal data, they exhibit limited generalization ability and insufficient matching stability. Their performance is particularly challenged in complex mountainous environments with repetitive textures and is often constrained by the robustness of feature extraction.

\subsection{Learning-based Methods}
In recent years, deep learning-based image matching methods have garnered significant attention. Notable advancements have been made in matching techniques for homologous natural images~\cite{mishchuk2017working,tian2020hynet,tian2019sosnet,liu2019gift}, with approaches such as SuperGlue~\cite{2019SuperGlue} and LightGlue~\cite{10377620} leveraging graph neural networks to learn robust descriptors, substantially enhancing matching performance. LoFTR~\cite{sun2021loftr} further introduces a Transformer-based detectorless approach that performs well in weakly textured regions, while RoMa~\cite{edstedt2024roma} develops a robust and dense matching model that achieves state-of-the-art results on real-world data. However, these methods are primarily trained on homologous natural image datasets (e.g., HPatches~\cite{8099893} and ScanNet~\cite{8099744}) and are not optimized for multimodal data, making them less effective in handling NRDs in multimodal image matching. To address this challenge, researchers have proposed data-driven multimodal image matching methods such as ReDFeat~\cite{9999700}, XoFTR~\cite{tuzcuouglu2024xoftr}, and GRiD~\cite{10715536}, which train networks on paired data to improve matching robustness. While these approaches demonstrate promising results on specific datasets, their generalization ability remains limited due to constraints in modal feature extraction. Consequently, their matching accuracy degrades significantly on unseen domains, making it difficult to adapt to complex cross-region and cross-dataset matching scenarios.

Unlike these methods, our proposed PromptMID requires training on only a single dataset, leveraging pre-trained diffusion models and VFMs to learn modality-invariant features. It demonstrates strong generalization across regions and datasets.

\subsection{Applications of foundation model}
In recent years, foundational models have driven a new wave of advancements in artificial intelligence. The availability of massive training data has endowed these models with strong zero-shot generalization capabilities. Models such as GPT-4~\cite{achiam2023gpt}, SAM~\cite{kirillov2023segany}, DINOv2~\cite{oquab2023dinov2}, CLIP~\cite{radford2021learning}, and Diffusion~\cite{SD} have significantly accelerated progress across various domains. Researchers have increasingly explored the applicability of foundational models in diverse fields. In remote sensing, CRS-Diff~\cite{10663449} has investigated remote sensing image generation using diffusion models, producing high-quality training data for downstream tasks. SGDM~\cite{WANG2025125} introduced a novel super-resolution paradigm for remote sensing images based on diffusion models, demonstrating superior performance. Additionally, RSPrompter~\cite{10409216} and SAM-RSIS~\cite{10680168} proposed remote sensing image instance segmentation methods leveraging the visual foundational model SAM, achieving remarkable results. The main objective of ours study is to adapt foundation models from the natural image domain to optical and SAR image matching, bridging the applicability gap between natural and remote sensing image domains. Our proposed PromptMID utilizes text prompts based on land use classification as priors information, along with a pre-trained diffusion models and the visual foundational models, to extract modality-invariant features and generate more robust descriptors.

\begin{figure*}[!t]
	\centering
	\includegraphics[width=0.95\linewidth]{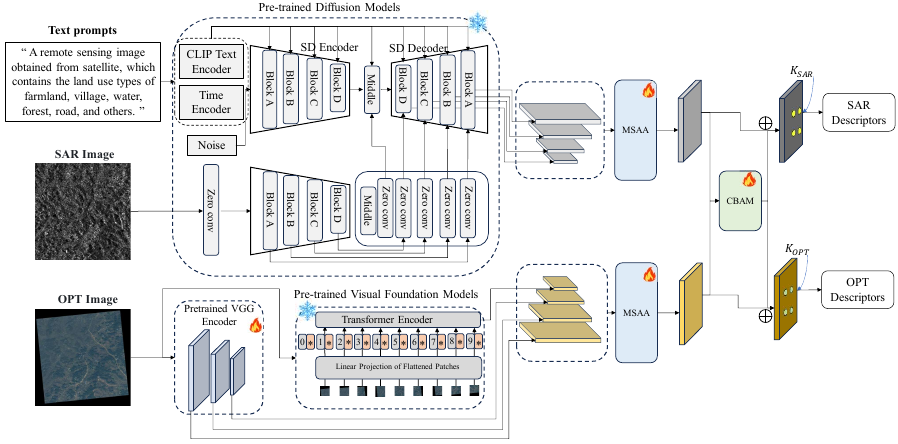}
	\caption{The flowchart of our proposed PromptMID is as follows: Initially, multi-scale features are extracted using pre-trained diffusion models and VFMs. These features are then aggregated at different scales through the MSAA module to fuse information at varying granularities. Finally, to alleviate the information redundancy introduced by multi-scale feature fusion, the CBAM module is applied in both spatial and channel dimensions, effectively suppresses irrelevant feature interference while enhancing feature representation.}
	\label{fig.all}
\end{figure*}

\section{Method}
\label{3}
\subsection{Main idea}
The main idea of our proposed PromptMID is shown in Fig.~\ref{fig.ggg} (e). Existing learning-based optical-SAR image descriptors usually rely on encoder variants to extract modal invariant features, as shown in Fig.~\ref{fig.ggg} (a)-(d). Its mathematical expression form is as follows:
\begin{equation}
\begin{aligned}
D^o &= \{d_j^o = f_{\theta}(p_j^o,OPT)\}_{j=1}^{M}, \\
D^s &= \{d_i^s = f_{\theta}(p_i^s,SAR)\}_{i=1}^{N}.
\end{aligned}
\end{equation}
Where $D^o$ and $D^s$ represent the optical and SAR descriptors, respectively, while $p_j^o$ and $p_i^s$ denote their corresponding sets of keypoints. Matching constraints are designed to minimize the descriptor distances of matched point pairs, while maximizing the distances between unmatched pairs. While this approach performs well within the seen domain, its limited generalization ability to unseen domains poses a significant challenge for practical applications. Specifically, during generalization in the unseen domain, the textural and semantic ambiguity of SAR images can hinder the clear distinction of land use categories. This confusion in extracting modality-invariant features impairs the similarity between SAR and optical image features, reducing matching accuracy. For instance, distinguishing between farmland and water bodies in SAR images may require geographic a priori information, as illustrated in Fig.~\ref{fig.question}.

Considering the above issues and drawing inspiration from foundation models, we propose a novel method, PromptMID, to enhance generalization in unseen domains, as shown in Fig.~\ref{fig.ggg} (e). Its mathematical expression is as follows:
\begin{equation}
\begin{aligned}
D_o &= \{d_j^o = MSAA(VFMs(p_j^o,OPT))\}_{j=1}^{M}, \\
D_s &= \{d_i^s = MSAA(D(p_i^s,SAR,Prompts))\}_{i=1}^{N}.
\end{aligned}
\end{equation}
where $D$ and $VFMs$ represent pre-trained diffusion models and visual foundation models, respectively; $MSAA$ denotes multi-scale aware aggregation module and $Prompts$ denotes text prompts based on land use classification as priors information. This method allows the a priori information to better guide modality-invariant feature extraction, ensures the semantic consistency of the features, and significantly improves the generalization in the unseen domain.

The detailed workflow is illustrated in Fig.~\ref{fig.all}. In the following subsections, we introduce the pre-trained diffusion models, pre-trained visual foundation models, the MSAA module, and the loss function.

\begin{figure*}[!t]
	\centering
	\includegraphics[width=0.8\linewidth]{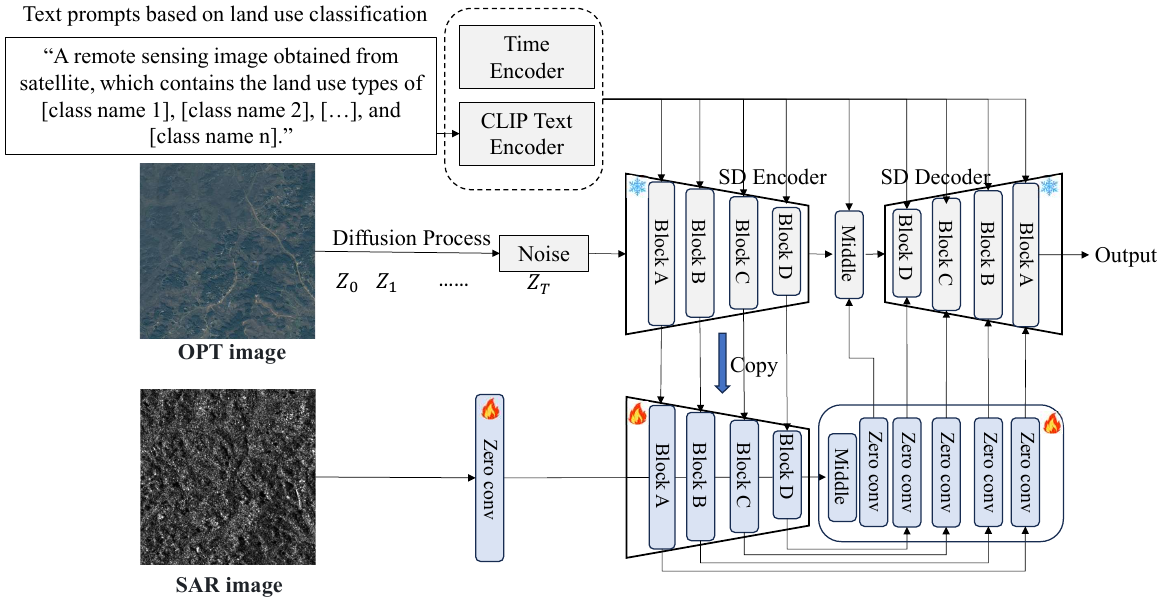}
	\caption{The flowchart of the pre-trained diffusion model illustrates the process. During the forward diffusion stage, Gaussian noise is progressively added to the optical image until it fully degrades into pure noise. In the reverse denoising stage, the SAR image, combined with textual cues, serves as a conditional input to guide the diffusion model in iteratively estimating and removing noise, ultimately reconstructing a clear optical image. The snowflake icon in the figure indicates parameter freezing, and the fire symbol indicates parameter training.}
	\label{fig.SD}
\end{figure*}

\subsection{Pre-trained diffusion models}
The workflow of the pre-trained diffusion model is shown in Fig.~\ref{fig.SD}. The diffusion model consists of a forward diffusion process, which progressively adds noise, and reverse denoising process, which gradually removes noise. The mathematical formulation of the forward diffusion process is given by:
\begin{equation}
q(x_t | x_{t-1}) = \mathcal{N}(x_t; \sqrt{1 - \beta_t} x_{t-1}, \beta_t I)
\end{equation}
where \( x_t \) represents the noisy data at time step \( t \), and \( x_{t-1} \) denotes the data from the previous step. The term \( \beta_t \) controls the variance of the added noise, while \( I \) is the identity matrix. The time step \( t \) ranges from \( t \in \{0, 1, 2, \dots, T\} \), and when \( t = T \), the data \( x_T \) approaches pure Gaussian noise after \( T \) iterations of noise addition. In the reverse denoising process, the goal is to iteratively denoise the random noise \( x_T \) step by step, starting from a standard Gaussian distribution. The mathematical expression for this process is shown below:
\begin{equation}
    p_\theta(x_{t-1} | x_t) = \mathcal{N}(x_{t-1}; \mu_\theta(x_t, t), \Sigma_\theta(x_t, t))
\end{equation}
where $ \theta $ is the network parameters obtained through our training. The random noise is progressively refined during the denoising process, ultimately yielding an approximation of the sample \( x_0 \). To control the generation of diffusion samples, a conditional diffusion model is introduced by incorporating conditional variables into the denoising process. The mathematical expression is as follows:
\begin{equation}
    x_{t-1} = \frac{1}{\sqrt{\alpha_t}} \left( x_t - \frac{1 - \alpha_t}{\sqrt{1 - \bar{\alpha}_t}} \epsilon_\theta(x_t, t, c) \right) + \sigma_t z
\end{equation}
where \( \epsilon_\theta(x_t, t, c) \) is the noise estimate after incorporating the condition \( c \), in addition to the current time step \( t \) and state \( x_t \). \( \sigma_t \) and \( z \) represent the random noise that can be added. Training diffusion models typically requires extensive computational resources and large datasets. Given the strong domain generalization capabilities of diffusion models, we use ControlNet~\cite{zhang2023adding} fine-tuning settings, using SAR images as conditions and optical images as targets. Additionally, text prompts are generated based on land use classification, with semantic information describing SAR images serving as priors information to guide the extraction of modality-invariant features and the details of the construction of the text prompts shown in Fig.~\ref{datasets_img}.

To address the challenges of unknown domain generalization caused by texture and semantic ambiguities in SAR images, as illustrated in Fig.~\ref{fig.question}. We propose a textual cue-guided pre-trained diffusion model based on land-use classification. The overall framework is depicted in Fig.~\ref{fig.SD}. Specifically, we freeze the parameters of the original Stable Diffusion (SD) model to preserve its generalization knowledge learned from large-scale image data. We create a trainable copy of the SD model and fine-tune it on our dataset, effectively mitigating the domain generalization problem. In the second stage of descriptor learning, we extract the decoder features from the pre-trained diffusion models as potential feature mappings, which are then used for descriptor contrastive learning alongside the features of the VFMs, as shown in Fig.~\ref{fig.all}.

\subsection{Visual foundation models}
With advancements in visual foundation models, such as GPT-4~\cite{achiam2023gpt}, SAM~\cite{kirillov2023segany}, DINOv2~\cite{oquab2023dinov2}, CLIP~\cite{radford2021learning}, the ability to extract features with strong domain generalization has significantly improved. In this paper, we choose DINOv2 as the feature extractor due to its training on billions of images, its strong domain generalization capabilities, and its proven excellent performance in various downstream tasks as a general-purpose visual representation. 

The DINOv2 pre-trained models excel at extracting high-level representations from images, but they have limitations in extracting detailed features, which can hinder precise localization. To address this, we introduce the VGG model pre-trained on ImageNet to capture finer details, such as boundaries and local features, that are crucial for accurate matching. By combining the coarse features from DINOv2 with the fine features from VGG19, we enable a more comprehensive multi-scale analysis of the image. This approach improves image matching accuracy while retaining the domain generalization advantages of the foundation model, as shown in Fig.~\ref{fig.all}.

\begin{figure}[!t]
	\centering
	\includegraphics[width=\linewidth]{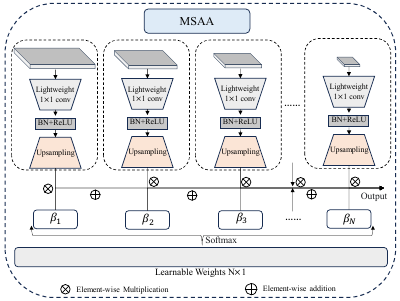}
	\caption{Our proposed MSAA module structure, integrates features across different scales and fuses information at various granularities.}
	\label{fig.msaa}
\end{figure}

\subsection{Feature aggregation}
After pre-trained the diffusion models and VFMs to obtain multi-scale modal-invariant features, we introduce a Multi-Scale Aware Aggregation (MSAA) module to enhance the model ability to comprehend complex scenes, as shown in Fig.~\ref{fig.msaa}. MSAA efficiently fuses global and local information, strengthening the model representational capacity across different scales. Lower-level features capture fine details, while higher-level features capture semantics. Simple concatenation of these features may lead to information distortion. To address this, MSAA first constructs multiple independent heads to process features at different scales and upsamples them to a unified scale. This process can be formulated as:
\begin{align}
F_i &\in \mathbb{R}^{C \times H_i \times W_i}, \quad i = 1, 2, \dots, N \\
F_i' &= \mathit{Upsample}(\mathit{Conv}(F_i), (H_i, W_i))
\end{align}
where $F_i$ is the feature map at the $i$ scale, $C$ is the number of channels, $H_i$ and $W_i$ is the spatial dimension of the feature map at the $i$ scale.
\begin{align}
W &= [w_1, w_2, \dots, w_N] \in \mathbb{R}^{N \times 1}\\
\beta_i &= \frac{\exp(w_i)}{\sum_{j=1}^{N} \exp(w_j)}, \quad i \in \{1,2,\dots,N\}\\
F^{'} &= \sum_{i=1}^{N} \beta_i \cdot F_i
\end{align}
where \(w_i\) is a learnable parameter for each scale, optimized via backpropagation, and \(\beta_i\) represents the result of Softmax normalization applied to the learnable weights \(w_i\). $F^{'}$ denotes the aggregated feature map.

MSAA effectively integrates features of different scales and dynamically adjusts the importance of these features by automatically sensing each scale's contribution to the results through learnable weights. However, after multi-scale feature fusion, the resulting features may contain redundant information. To address this, we introduce the CBAM~\cite{cbam} attention mechanism to filter out the most representative features, reducing the interference of irrelevant ones. Enhance key channel features using channel attention and emphasize crucial areas through spatial attention. This improves the accuracy of the final features and enhances the discriminative ability when constructing descriptors.

\subsection{Loss function}
The training process of our proposed PromptMID consists of two phases. The first phase involves training the pre-trained diffusion model, as depicted in Fig.~\ref{fig.SD}. The second phase corresponds to training the descriptor learning network, as shown in Fig.~\ref{fig.all}.

The flowchart of the first stage is shown in Fig.~\ref{fig.SD}. The trainable parameters are represented using the fire symbol, and the loss function is as follows:
\begin{equation}
L=\mathbb{E}_{\left(z_0, \epsilon,c, t\right)} \| \epsilon-\epsilon_\theta\left(z_t, t, c\right) \|^2
\end{equation}
where \( \epsilon \) denotes Gaussian noise, time step \( t \) ranges from \( t \in \{0, 1, 2, \dots, T\} \) and \( c \) represents the input condition, which includes the Prompts and SAR images.

The flowchart of the second stage is shown in Fig.~\ref{fig.all}. Upon completing the first stage of training and obtaining the pre-trained diffusion model, we extract multi-scale features from its decoder and perform descriptor learning by combining these features with those extracted from the VFMs. The trainable parameters are represented using the fire symbol, and the loss function is as follows:
\begin{equation}
L(D_{M}^{o},D_{N}^{s}) = \sum_{s}^{n} -log\frac{exp(sim(d_i^{o},d_i^{s})/\tau)}{\sum_{k=1}^{n}exp(sim(d_i^{o},d_i^{s})/\tau)}
\end{equation}
where $i$ denotes any descriptor within the batch. $sim$ denotes cosine similarity, $\tau$ denotes softmax temperature.

\section{Experiments}
\label{4}
In this section, we present the experimental setup, dataset, and results. Our PromptMID was compared with twelve algorithms: SIFT~\cite{lowe2004distinctive}, RIFT~\cite{8935498}, LNIFT~\cite{li2022lnift}, SRIF~\cite{li2023multimodal}, LoFTR~\cite{sun2021loftr}, XoFTR~\cite{tuzcuouglu2024xoftr}, RoMa~\cite{edstedt2024roma}, ReDFeat~\cite{9999700}, HardNet~\cite{mishchuk2017working}, HyNet~\cite{tian2020hynet}, SoSNet~\cite{tian2019sosnet} and GIFT~\cite{liu2019gift}. Finally, we performed ablation study to validate the effectiveness of the key components of the PromptMID.

\subsection{Implementation details}
\label{indetails}
During the first stage of the diffusion fine-tuning process, SAR images were used as the conditional input, with prompt text based on land use classification constructed as an additional conditional input, while optical images served as the ground truth labels. The pre-trained diffusion model for the first stage of pre-training was the Stable Diffusion V1.5 weights, optimized using AdamW with a batch size of 2, a learning rate of $1 \times 10^{-5}$, and a weight decay of 0.01, trained for 50 epochs. During the second stage of the descriptor learning process, the optical images were randomly rotated within a range of [-10, 10] degrees and subjected to random scaling within the range of [0.8, 1], while the SAR images utilized pre-trained diffusion model weights from the first stage to extract modality-invariant features. The descriptor learning model is optimized using AdamW with a batch size of 1, a learning rate of $1 \times 10^{-4}$, and a weight decay of 0.01, trained for 20 epochs ,with the learning rate decreasing progressively over epochs.

\begin{figure}[!t]
	\centering
	\includegraphics[width=\linewidth]{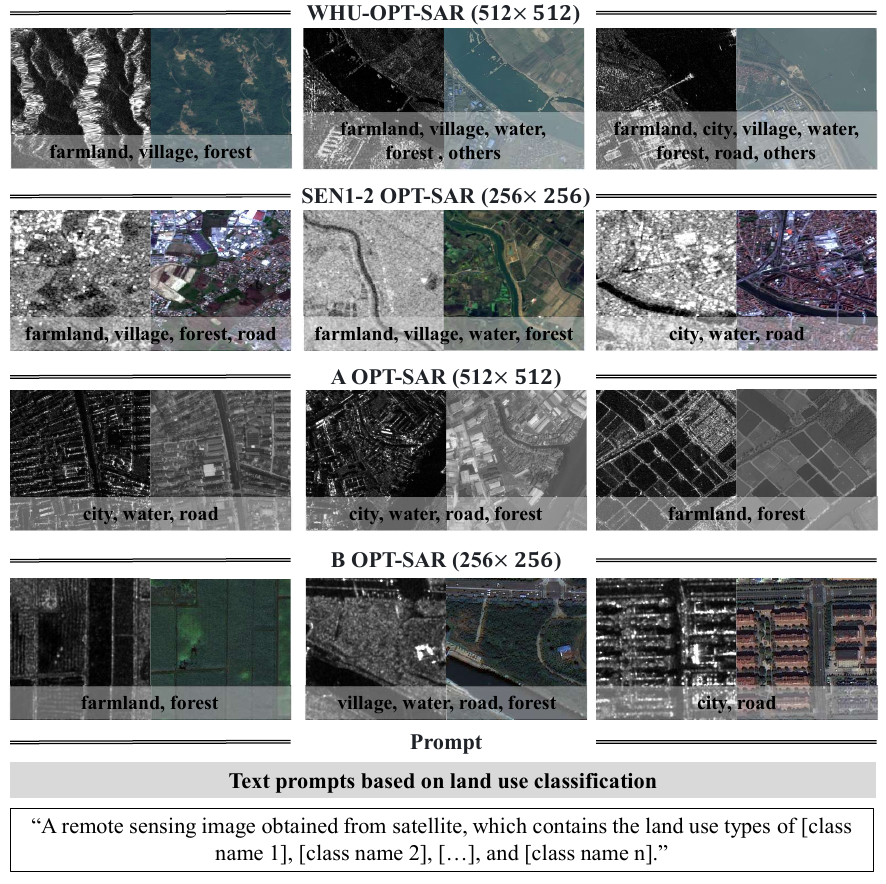}
	\caption{The four datasets and their corresponding land use classifications are as follows. WHU-OPT-SAR~\cite{LI2022102638} is the dataset used for training. The other three datasets are used to evaluate its zero-shot generalizability. Additionally, text prompts were constructed for each dataset based on its land use classification.}
	\label{datasets_img}
\end{figure}

\begin{table*}[ht]
    \centering
    \caption{Comparison of evaluation metrics for different methods in seen domains.}
    \label{tab:seen}
    \renewcommand{\arraystretch}{1.2}
    \begin{tabular}{lcccccc} 
        \toprule
        \multicolumn{4}{c}{} & \multicolumn{3}{c}{\textbf{Seen Domain (WHU-OPT-SAR)}} \\
        \cmidrule(lr){5-7}
        Category & Methods & Multimodal & Reference & SR(\%)$\uparrow$ & NCM$\uparrow$ & RMSE$\downarrow$  \\
        \midrule
        \multirow{4}{*}{Traditional} 
        & SIFT~\cite{lowe2004distinctive}  & \xmark & IJCV'04 & 0.5 & 11 & \textbf{1.310} \\
        & RIFT~\cite{8935498} & \cmark & TIP'20 & 22.5 & 85 & 2.027 \\
        & LNIFT~\cite{li2022lnift} & \cmark & TGRS'22 & 40.0 & 27 & 2.084 \\
        & SRIF~\cite{li2023multimodal} & \cmark & ISPRS'23 & 72.0 & 61 & 2.075 \\
        \midrule
        \multirow{3}{*}{Detector-free} 
        & LoFTR~\cite{sun2021loftr} & \xmark & CVPR'21 & 39.5 & 62 & 1.809 \\
        & XoFTR~\cite{tuzcuouglu2024xoftr} & \cmark & CVPR'24 & 24.0 & 51 & 1.843 \\
        & RoMa~\cite{edstedt2024roma} & \xmark & CVPR'24 & 52.0 & \underline{2835} & \underline{1.802} \\
        \midrule
        \multirow{1}{*}[1.2ex]{\makecell{Joint Detection \\ and Description}} 
        & \rule{0pt}{3ex} ReDFeat~\cite{9999700} & \cmark & TIP'23 & \underline{88.0} & 262 & 1.896 \\[1.5ex]
        \midrule
        \multirow{5}{*}{Detect-then-Describe} 
        & HardNet~\cite{mishchuk2017working} & \xmark & NeurIPS'17 & 38.0 & 34 & 1.799 \\
        & HyNet~\cite{tian2020hynet} & \xmark & NeurIPS'20 & 4.5 & 20 & 1.813 \\
        & SosNet~\cite{tian2019sosnet} & \xmark & CVPR'19 & 7.5 & 17 & 1.942 \\
        & GIFT~\cite{liu2019gift} & \xmark & NeurIPS'19 & 54.5 & 37 & 1.892 \\
        \cmidrule(lr){2-7} 
        & \textbf{PromptMID (Ours)}  & \cmark & - & \textbf{100} & \textbf{452} & 1.870 \\
        \bottomrule
    \end{tabular}
\end{table*}

\subsection{Datasets}
\label{Datasets}
In order to accurately assess the performance of the proposed model, we proposed a cross-dataset evaluation benchmark in which the model was trained on an independent dataset and tested on four previously unseen domains. Referring to the setup of paper~\cite{LI2022102638}, we adopted the Chinese Land Use Classification Standard (GB/T 21010-2017) and consolidated the land use categories into seven classes: farmland, city, village, water, forest, road, and others.

\subsubsection{Training Datasets}
We employed the WHU-OPT-SAR dataset~\cite{LI2022102638} as the training dataset, which contains 100 optical and SAR image pairs of the same geographic region. The optical images were acquired by the Chinese Gaofen-1 (GF-1) satellites with a ground resolution of 5 meters, while the SAR images were obtained from the Chinese Gaofen-3 (GF-3) satellites. The dataset spans over 50,000 square kilometers in Hubei Province, China, encompassing diverse terrain types and vegetation, with pixel-level land use classification labels. Each optical image consists of four channels: red (R), green (G), blue (B), and near-infrared (NIR). The optical and SAR images, originally sized at 5556 × 3704 pixels, were cropped into non-overlapping 512 × 512 pixel image blocks after eliminating the NIR channel, resulting in a total of 7000 image pairs. Among these, 6800 pairs were randomly selected for the training set, and 200 pairs were designated as the test set. As shown in Fig.~\ref{datasets_img} first line.

\subsubsection{Test Datasets}
In order to evaluate the cross-domain generalization capability of the proposed method, we use three independent unseen domains and one seen domains for comprehensive evaluation. The details are as follows.

\textit{\textbf{(1) SEN1-2 OPT-SAR}:} The Sentinel-1 and Sentinel-2 (SEN1-2) datasets serve as the first unseen OPT-SAR datasets utilized in this study. This dataset contains 282,384 optical and SAR image blocks of size 256 × 256 pixels, consisting of multi-seasonal images from all over the world. These images include a variety of land covers, such as mountains, towns, rivers, farmlands, and more, from which we randomly selected 100 pairs for testing. Compared to the training dataset WHU-OPT-SAR~\cite{LI2022102638}, there is a significant domain gap, making the matching task more challenging. This is illustrated in Fig.~\ref{datasets_img} second line.

\textit{\textbf{(2) A OPT-SAR}:} The second unseen OPT-SAR dataset~\cite{9204802} consists of SAR imagery acquired by Chinese Gaofen-3 (GF-3) and optical imagery of the corresponding area captured from Google Earth at a resolution of 1 meter. We follow the experimental setup described in the paper~\cite{9999700} and use its 424 pairs of optical and SAR images as the test dataset. Compared to the training dataset WHU-OPT-SAR~\cite{LI2022102638}, this dataset exhibits cross-region differences, and the matching difficulty is relatively low. This is illustrated in Fig.~\ref{datasets_img} third line.

\textit{\textbf{(3) B OPT-SAR}:} The third unseen OPT-SAR dataset~\cite{REN2022102896} consists of SAR imagery captured by Chinese Gaofen-3 (GF-3) and optical imagery captured by Chinese Gaofen-2 (GF-2) at a resolution of 1 meter. We randomly selected 200 pairs of optical and SAR datasets from Dongying city in Shandong Province in China as test data. This is illustrated in Fig.~\ref{datasets_img} fourth line.

\subsection{Evaluation Metrics}
We follow the evaluation metrics specified in the referenced paper~\cite{li2023multimodal,10662912}, utilizing three quantitative metrics: the number of correct matches (NCM), root mean square error (RMSE), and matching success rate (SR). We first apply random rotations within \([-10, 10]\) degrees and random scale transformations within \([0.8, 1]\) to the optical images in the test dataset generate the ground truth homography matrix. If the NCM is less than 10, the match is considered a failure, the RMSE is set to 20, and it is excluded from the overall NCM and RMSE calculations.

\subsection{Experiments on seen Domains}
Table \ref{tab:seen} presents a comparison of different methods based on evaluation metrics on the Seen Domain (WHU-OPT-SAR) dataset, including SR, NCM, and RMSE. The matching success rate of SIFT is notably low at 0.5\%, highlighting its limitation in cross-modal matching tasks. This is primarily because SIFT relies heavily on texture and gradient information, which struggles to adapt to the substantial modality differences between optical and SAR images. RIFT demonstrates better performance by constructing MIM maps to extract modal-invariant features for multimodal matching, achieving an SR of 22.5\%. However, its RMSE remains high at 2.027. LNIFT improves the SR metric to 40.0\%, but its low NCM suggests unstable matching quality. SRIF further enhances SR to 72.0\%, yet the NCM remains suboptimal at 61, and the RMSE is still relatively high at 2.075, indicating that the feature descriptor stability requires further optimization. Overall, while methods such as RIFT, LNIFT, and SRIF introduce multimodal adaptations, they primarily rely on local feature extraction and matching. Consequently, they struggle to globally model the nonlinear mapping between optical and SAR images in complex mountainous regions. This results in unstable matching quality, as evidenced by significant fluctuations in NCM and large RMSE errors.

\begin{figure*}[h]
	\centering
	\includegraphics[width=0.8\linewidth]{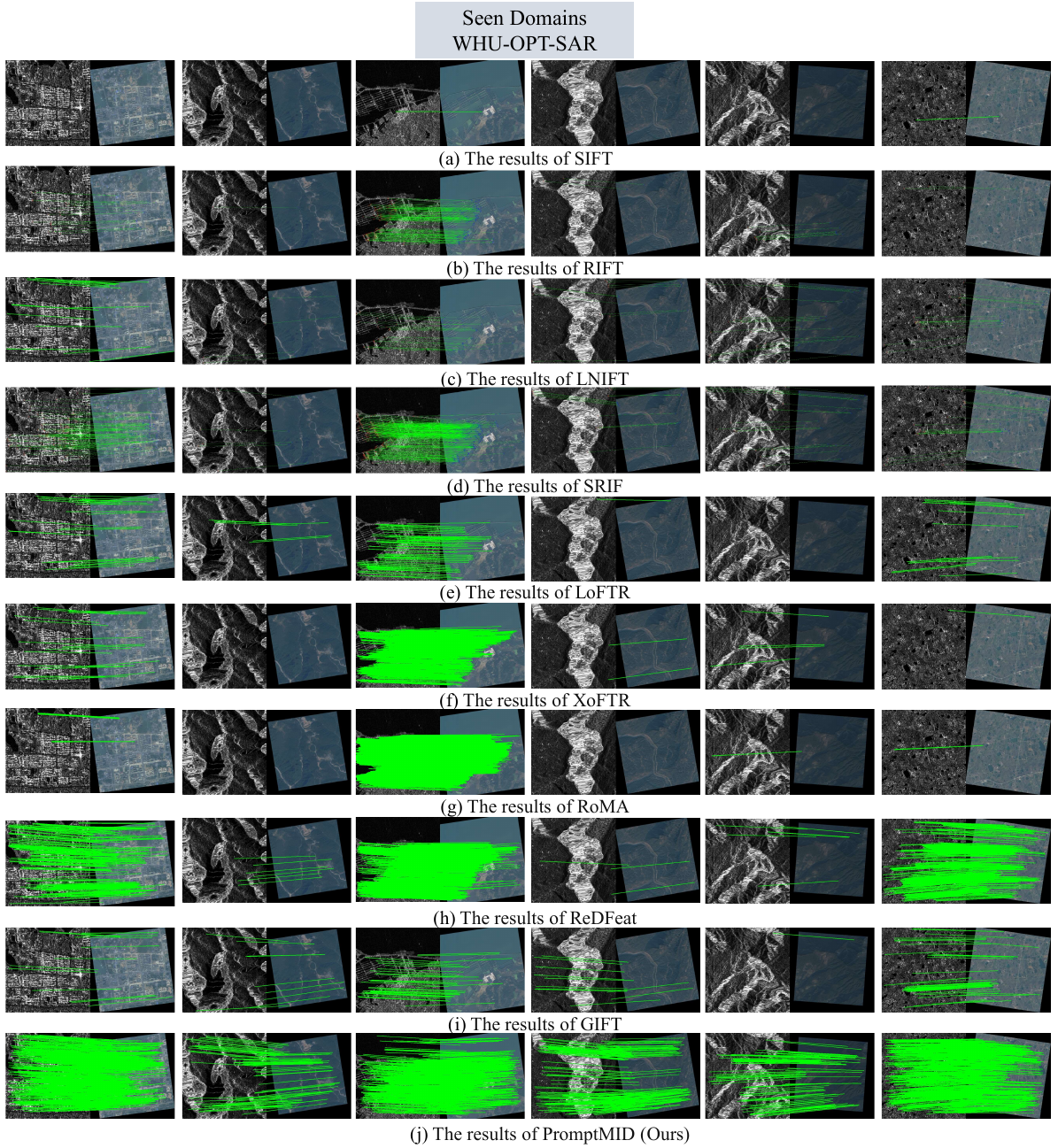}
	\caption{Qualitative comparison results of test methods on typical image pairs within the seen domain. We showcase matched image pairs with an RMSE below 3, displaying only successfully matched point pairs. The green connecting lines represent these successful matches.}
	\label{seen}
\end{figure*}

LoFTR adopts a detector-free matching method, but its matching performance on cross-modal data is limited, with an SR of only 39.5\% and an NCM of 62, indicating that it is still affected by modal differences in the feature matching stage. XoFTR, which is specifically optimised for multimodal matching, instead has a lower SR of 24.0\% and a lower NCM, probably due to its insufficient generalization ability in the optical-SAR task, probably due to its insufficient generalization ability. In contrast, RoMa incorporates VFMs and achieves a considerably higher SR of 52.0\%, further demonstrating the potential of foundational visual models for matching tasks. However, its NCM value is exceptionally high (2835), primarily due to the adoption of a dense matching strategy, which significantly increases the number of successfully matched point pairs. Despite these improvements, RoMa still falls short of the 100\% SR achieved by our proposed PromptMID in optical-SAR matching. This suggests that solely leveraging a foundational vision model trained on natural images is insufficient to fully address the challenges of cross-modal matching, highlighting the necessity of a more targeted adaptation strategy.

ReDFeat employs an end-to-end joint detection and description learning strategy, achieving superior results with an SR of 88.0\% and an NCM of 262. This indicates a high quality of matched points and highlights the effectiveness of joint learning compared to the detect-then-describe methods. In contrast, HardNet, HyNet, and SosNet perform poorly in the optical-SAR matching task, with SR values below 40\% and relatively low NCM scores. GIFT performs comparatively well, achieving an SR of 54.5\%, yet it still falls short of delivering the best matching performance.

Our proposed PromptMID achieves a perfect SR of 100\%, demonstrating its ability to perform stable cross-modal matching on the WHU-OPT-SAR dataset. Additionally, its NCM value (452) significantly surpasses that of existing methods, indicating a higher quality of matching points. This performance gain is primarily attributed to the combination of pre-trained diffusion models and VFMs for extracting modality-invariant features, while the MSAA module effectively aggregates multi-scale features to enhance the discriminative power of the descriptors. Figure~\ref{seen} illustrates the matching results across multiple methods. The second, fourth, and fifth image pairs depict complex mountainous scenes with severe speckle noise, where conventional methods (e.g., SIFT, RIFT, LNIFT, and SRIF) fail to produce accurate matches, whereas the proposed PromptMID achieves successful matching. Additionally, while dense matching methods such as LoFTR, XoFTR, and RoMA extract more matches in the third image pair, their generalization ability in optical-SAR image matching remains weaker than that of PromptMID. In contrast, ReDFeat demonstrates greater overall stability but experiences a significant reduction in matches under severe speckle noise. Collectively, these results underscore PromptMID as the most robust method, achieving the highest matching success rate across all image pairs.

\begin{table*}[h]
    \centering
    \caption{Comparison of evaluation metrics for different methods in unseen domains.}
    \label{tab:unseen}
    \renewcommand{\arraystretch}{1.2}
	\scalebox{1}{
    \begin{tabular}{lcccc|ccc|ccc}
    \toprule
        \multicolumn{2}{c}{} & \multicolumn{9}{c}{\textbf{Unseen Domain}} \\
        \cmidrule(lr){3-11} 
        \multicolumn{2}{c}{} & \multicolumn{3}{c}{\textbf{SEN1-2 OPT-SAR}} & \multicolumn{3}{c}{\textbf{A OPT-SAR}} & \multicolumn{3}{c}{\textbf{B OPT-SAR}}  \\
        \cmidrule(lr){3-5} \cmidrule(lr){6-8} \cmidrule(lr){9-11} 
        Category & Methods  
        & SR(\%)$\uparrow$ & NCM$\uparrow$ & RMSE$\downarrow$  
        & SR(\%)$\uparrow$ & NCM$\uparrow$ & RMSE$\downarrow$  
        & SR(\%)$\uparrow$ & NCM$\uparrow$ & RMSE$\downarrow$    \\
        \midrule
        \multirow{4}{*}{Traditional} 
        & SIFT~\cite{lowe2004distinctive}  & 0.0 & — & —  & 0.0 & — & —  & 0.0 & — & —  \\
        & RIFT~\cite{8935498}  & 63.0 & 49 & 2.078  & 78.8 & 63 & 2.071  & 72.0 & 35 & 2.116   \\
        & LNIFT~\cite{li2022lnift}  & 97.5 & 174 & 2.111  & 90.6 & 51 & 2.101  & 95.5 & 154 & 2.100  \\
        & SRIF~\cite{li2023multimodal}  & 98.5 & \underline{304} & 2.108  & \underline{98.8} & 153 & 2.097  & \textbf{97.0} & \textbf{229} & 2.140  \\
        \midrule
        \multirow{3}{*}{Detector-Free} 
        & LoFTR~\cite{sun2021loftr}  & 38.0 & 21 & 1.854 & 27.12 & 30 & 1.917  & 9.0 & 21 & 1.831 \\
        & XoFTR~\cite{tuzcuouglu2024xoftr}  & 20.0 & 24 & \underline{1.845}  & 76.4 & 111 & 1.868  & 26.0 & 31 & 1.921   \\
        & RoMa~\cite{edstedt2024roma}  & 41.0 & 42 & 2.047  & 79.2 & \textbf{1867} & \underline{1.915}  & 49.5 & \underline{163} & 1.980  \\
        \midrule
        \multirow{1}{*}[1.2ex]{\makecell{Joint Detection \\ and Description}}
        & \rule{0pt}{3ex} ReDFeat~\cite{9999700}  & \underline{98.8} & \textbf{322} & 1.870  & \textbf{100} & 268 & 1.919  & 83.5 & 148 & 1.997   \\[1.5ex]
        \midrule
        \multirow{5}{*}{Detect-Then-Describe} 
        & HardNet~\cite{mishchuk2017working}  & 73.0 & 25 & 1.860  & 37.7 & 18 & 1.875  & 17.5 & 19 & 1.948  \\
        & HyNet~\cite{tian2020hynet}  & 64.0 & 21 & \textbf{1.795}  & 7.8 & 15 & 1.859  & 18.0 & 17 & 1.894 \\
        & SosNet~\cite{tian2019sosnet}  & 67.0 & 21 & 1.808  & 12.0 & 14 & 1.868  & 18.0 & 17 & 1.894  \\
        & GIFT~\cite{liu2019gift}  & 71.0 & 30 & 1.922  & 76.9 & 40 & 1.923  & 32.0 & 23 & 1.954  \\
        \cmidrule(lr){2-11} 
        & \textbf{PromptMID (Ours)}  & \textbf{99.0} & 152 & 1.861  & \textbf{100} & \underline{471} & \textbf{1.841} & \underline{96.0} & 98 & \textbf{1.885} \\
        \bottomrule
    \end{tabular}}
\end{table*}

\begin{figure*}[h]
	\centering
	\includegraphics[width=0.8\linewidth]{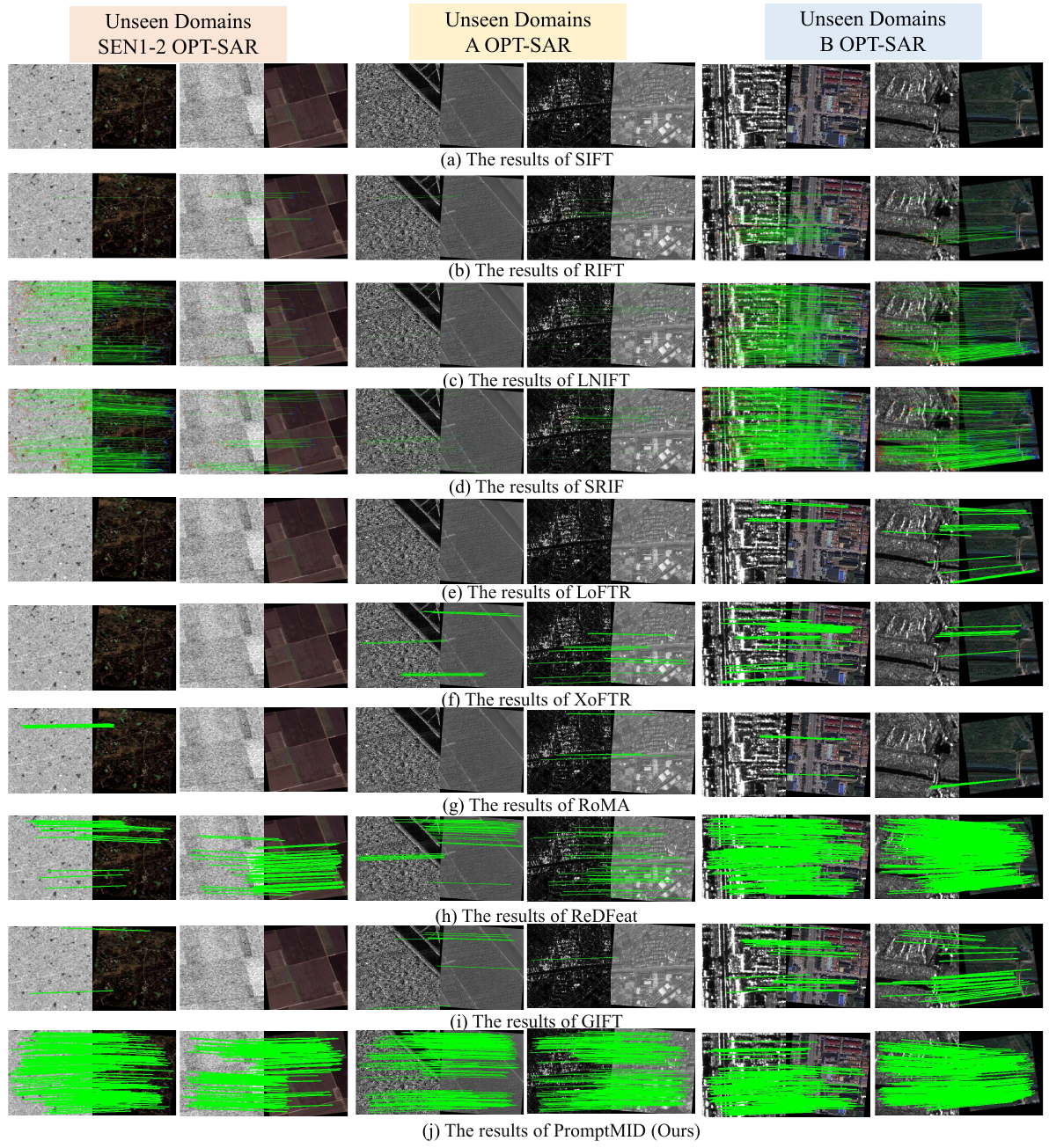}
	\caption{Qualitative comparison results of test methods on typical image pairs within the unseen domain. We showcase matched image pairs with an RMSE below 3, displaying only successfully matched point pairs. The green connecting lines represent these successful matches.}
	\label{unseen}
\end{figure*}

\subsection{Experiments on Unseen Domains}
Many matching methods perform well on seen domains but exhibit reduced generalization capability in unseen domains. To assess the cross-regional and cross-dataset generalization ability of our proposed PromptMID, we compare the evaluation metrics of various methods in unseen domains. This evaluation spans three distinct datasets: SEN1-2 OPT-SAR, A OPT-SAR, and B OPT-SAR, as shown in Table \ref{tab:unseen}.

SIFT fails completely in the cross-modal matching task, achieving a matching success rate of 0\% (SR = 0\%) across all datasets. RIFT demonstrates improved matching rates with SR of 63.0\%, 78.8\%, and 72.0\% on the three datasets, yet its NCM and RMSE performance remain suboptimal, indicating that the quality of matched point pairs is relatively low. LNIFT achieves significantly higher SR of 97.5\%, 90.6\%, and 95.5\% compared to RIFT. However, its NCM value is relatively low, suggesting that while the number of matched points is large, their quality remains insufficient. SRIF attains an SR of 97.0\% on the B OPT-SAR dataset and exhibits a much higher NCM than RIFT and LNIFT, demonstrating the high quality of its matched point pairs. Notably, SRIF achieves better matching accuracy on B OPT-SAR than on the WHU-OPT-SAR dataset. This discrepancy arises from the WHU-OPT-SAR dataset location in a mountainous region, which contains a large number of repetitive textures and complex scenes, significantly reducing the matching accuracy of traditional methods and highlighting their inherent limitations.

LoFTR achieves SR of 38.0\% and 27.12\% on the SEN1-2 OPT-SAR and A OPT-SAR datasets, respectively, but drops significantly to 9.0\% on the B OPT-SAR dataset, indicating weak adaptability in cross-modal matching tasks. XoFTR attains its highest SR of 76.4\% on the A OPT-SAR dataset; however, its poor performance on other datasets highlights its limited generalization ability. RoMa achieves the highest NCM of 1867 on the A OPT-SAR dataset, significantly surpassing other methods, demonstrating its ability to identify high-quality matches in specific scenarios. However, its SR on the B OPT-SAR dataset is only 49.5\%, revealing instability in matching performance. Since RoMa employs a dense matching strategy, the number of matched pairs is high, leading to a large NCM value. However, this does not guarantee stable matching performance across all scenarios.

ReDFeat achieves a 100\% SR on the A OPT-SAR dataset, with SR of 98.8\% and 83.5\% on the SEN1-2 OPT-SAR and B OPT-SAR datasets, respectively, demonstrating strong overall performance with a high NCM. However, its lower SR on the B OPT-SAR dataset suggests a decline in generalization performance due to significant scene variations. In contrast, HardNet, HyNet, and SosNet exhibit consistently lower matching rates across all datasets, with HyNet achieving only 18.0\% SR on B OPT-SAR, indicating significant challenges in cross-modal matching tasks. The GIFT method achieves 76.9\% SR on the A OPT-SAR dataset, but its overall NCM and RMSE metrics remain suboptimal.

Our proposed PromptMID outperforms all other methods, achieving SR of 99.0\% and 100\% on the SEN1-2 OPT-SAR and A OPT-SAR datasets, respectively, and 96.0\% on the B OPT-SAR dataset, demonstrating the highest overall matching success rate. Notably, PromptMID attains an NCM of 471 on the A OPT-SAR dataset, significantly surpassing other approaches, indicating the superior quality of its matched pairs. Moreover, the RMSE values remain the lowest across all datasets (1.841, 1.861, 1.885), further validating the high accuracy and robustness of PromptMID in cross-modal matching tasks within unseen domains. Figure ~\ref{unseen} presents a visualization of the matching results obtained by various methods, where only successfully matched point pairs are displayed. Green connecting lines indicate these successful matches. In two complex scenarios, SEN1-2 OPT-SAR and A OPT-SAR, conventional methods struggle to achieve accurate matching, particularly due to significant speckle noise in optical-SAR modalities. In these cases, the number of successfully matched points is extremely low or even nonexistent. In the B OPT-SAR scenario, certain methods (e.g., SIFT and LNIFT) can detect some matches, but their overall stability is poor, and the number of matches remains significantly lower than that of deep learning-based approaches. Among deep learning methods, LoFTR, XoFTR, and RoMA extract a substantial number of matches. However, their generalization ability is limited, and they produce matching errors even when differences between optical-SAR images are relatively minor. While ReDFeat demonstrates more consistent performance across most scenarios, its number of matching points drops sharply under severe speckle noise, indicating persistent challenges in high-noise environments. In contrast, PromptMID detects the highest density of matching points across all scenarios and achieves successful matching even in complex cases where both traditional and deep learning methods fail. Notably, under high-noise conditions (e.g., the fourth and fifth columns), PromptMID maintains a high matching success rate and exhibits the strongest generalization ability for unseen domains.

\begin{table*}[h]
    \centering
    \caption{Ablation experiments on Seen and Unseen Domains with different PromptMID component combinations. Here, \textbf{D} denotes the pre-trained diffusion models, \textbf{V} denotes the visual foundation models, and \textbf{F} refers to the feature aggregation module.}
    \renewcommand{\arraystretch}{1.3}
    \label{tab:ablation}
    \scalebox{1}{
    \begin{tabular}{l|c|c|c|ccc|ccc|ccc|ccc}
        \hline
        \multirow{3}{*}{Pipeline} & \multirow{3}{*}{D} & \multirow{3}{*}{V} & \multirow{3}{*}{F} 
        & \multicolumn{3}{c|}{Seen Domain} & \multicolumn{9}{c}{Unseen Domain} \\
        \cline{5-16}
        &  &  &  & \multicolumn{3}{c|}{WHU-OPT-SAR} & \multicolumn{3}{c|}{SEN1-2 OPT-SAR} & \multicolumn{3}{c|}{A OPT-SAR} & \multicolumn{3}{c}{B OPT-SAR} \\
        \cline{5-16}
        &  &  &  & SR(\%) & NCM & RMSE & SR(\%) & NCM & RMSE & SR(\%) & NCM & RMSE & SR(\%) & NCM & RMSE \\
        \hline
         Baseline &  &  &  & \textbf{100}  & \textbf{470} & \underline{1.872}  & 1.0    & 13  & 1.861  & \underline{99.76} & 399 & 1.903  & 0    & \textemdash  & \textemdash  \\
         V1 & \checkmark &  &  & \textbf{100}  & 374 & 1.883  & \underline{97.0}   & 101 & \textbf{1.833}  & \textbf{100}  & 387 & \underline{1.870}  & \underline{87.5} & 64  & \textbf{1.845}  \\
         V2 &  &\checkmark  &  & \textbf{100}  & 446 & 1.884  & 2.0    & 12  & 1.855  & 99.06 & 353 & 1.916  & 0.5  & 16  & 2.145  \\
         V3 &\checkmark  & \checkmark &  & \textbf{100}  & 409 & 1.888  & \textbf{99}   & \underline{109} & \underline{1.834}  & \textbf{100}  & 396 & 1.888  & \underline{87.5} & 59  & 1.902  \\
         V4 &\checkmark  &\checkmark  & \checkmark  & \textbf{100}  & \underline{452} & \textbf{1.870}  & \textbf{99}   & \textbf{152} & 1.861  & \textbf{100}  & \textbf{471} & \textbf{1.841}  & \textbf{96.0}   & \textbf{98}  & \underline{1.885}  \\
        \hline
    \end{tabular}}
\end{table*}

\subsection{Ablation experiments}
\label{xr}
\subsubsection{Effectiveness of PromptMID key components}
This ablation experiment evaluates different component combinations within the PromptMID framework to analyze the contribution of each module to cross-modal matching performance and assess their impact on both Seen and Unseen Domains, as shown in Table \ref{tab:ablation}. Specifically, D represents the pre-trained diffusion model, which generates prior features for cross-modal alignment, V denotes the visual foundation model, responsible for extracting high-level feature representations, and F refers to the feature aggregation module, which fuses multi-scale features. \textbf{Baseline} model (without D, V, and F) achieves a high success rate (SR=100\%) on the WHU-OPT-SAR dataset (Seen Domain). However, its performance significantly deteriorates on Unseen Domain datasets (SEN1-2 OPT-SAR, A OPT-SAR, and B OPT-SAR), where its generalization capability is notably poor. In particular, on the SEN1-2 OPT-SAR and B OPT-SAR datasets, the SR is nearly zero. This indicates that the Baseline method struggles to adapt effectively to unseen-domain data in cross-modal matching tasks, lacking sufficient domain generalization capability. \textbf{V1} contains only the D component, which significantly enhances the matching performance on the Unseen Domain datasets. Specifically, the SR on the SEN1-2 OPT-SAR and A OPT-SAR datasets reach 97.0\% and 100\%, respectively, which are substantially higher than those of the Baseline (1.0\% and 99.76\%). This demonstrates that the D component effectively preserves the modality invariance of descriptors and improves the generalization ability of the model in cross-modal matching tasks. However, on the B OPT-SAR dataset, the SR is only 87.5\%, indicating that relying solely on the D component still has limitations in complex scenarios. This suggests that using the diffusion model alone may struggle to ensure stable and robust matching in challenging conditions. \textbf{V2} contains only the V component, with SR of 2.0\% on the SEN1-2 OPT-SAR dataset and 0.5\% on the B OPT-SAR dataset in the Unseen Domain. These results show a significant decline compared to V1, indicating that the V component struggles to function independently in the cross-modal matching task. This limitation is likely due to the fact that the visual foundation model is primarily trained on unimodal data, making it difficult to handle optical-SAR matching, which involves substantial modality differences. \textbf{V3} integrates both the D and V components, achieving SRs of 99\%, 100\%, and 87.5\% on the SEN1-2 OPT-SAR, A OPT-SAR, and B OPT-SAR datasets in the Unseen Domain, respectively. These results demonstrate a notable improvement over V1 and V2, indicating that the combination of D and V effectively enhances cross-modal matching capability. However, the SR on the B OPT-SAR dataset remains at 87.5\%, suggesting that certain limitations persist in handling complex scenarios. \textbf{V4} integrates the D, V, and F components, forming the complete \textbf{PromptMID} model. Across all unseen domain datasets, the SR is significantly improved, particularly on B OPT-SAR, where it reaches 96.0\%, demonstrating a substantial enhancement compared to V1, V2, and V3. Additionally, the NCM and RMSE metrics are significantly improved, suggesting that the inclusion of the F-part further enhances the discriminative power of the descriptors while improving the stability and accuracy of the matching process. The method also maintains an SR of 100\% on WHU-OPT-SAR, underscoring its consistency across both Seen and Unseen domain tasks.

\begin{table*}[h]
    \centering
    \caption{Performance comparison of keypoint detection methods with PromptMID descriptor across four datasets.}
    \label{tab:keypoint_methods}
    \renewcommand{\arraystretch}{1.3}
    \setlength{\tabcolsep}{6pt}
    \begin{tabular}{l|ccc|ccc|ccc|ccc}
        \hline
        \multirow{2}{*}{Keypoint Method} & \multicolumn{3}{c|}{WHU-OPT-SAR} & \multicolumn{3}{c|}{SEN1-2 OPT-SAR} & \multicolumn{3}{c|}{A OPT-SAR} & \multicolumn{3}{c}{B OPT-SAR} \\
        \cline{2-13}
        & SR(\%)$\uparrow$ & NCM$\uparrow$ & RMSE$\downarrow$ & SR(\%)$\uparrow$ & NCM$\uparrow$ & RMSE$\downarrow$ & SR(\%)$\uparrow$ & NCM$\uparrow$ & RMSE$\downarrow$ & SR(\%)$\uparrow$ & NCM$\uparrow$ & RMSE$\downarrow$ \\
        \hline
        SIFT   & 75.5  & \underline{106} & \underline{1.873}  & \underline{91.0}  & 46  & \underline{1.843}  & \textbf{100} & 131 & 1.910  & \underline{42.5}  & 25  & 1.911  \\
        KeyNet & \underline{90.5}  & 33  & 1.948  & 70.0  & 21  & 1.888  & 96.93 & 35  & 1.941  & 34.0  & 22  & \underline{1.877}  \\
        SuperPoint & 73.5  & 49  & 1.947  & 16.0  & 19  & 1.861  & 61.79 & 24  & 1.977  & 2.5   & 24  & 1.791  \\
        GIFT   & 73.5  & 49  & 1.958  & 16.0  & 19  & 1.850  & 61.79 & 24  & 1.977  & 2.5   & 24  & 1.782  \\
        PC-FAST (Ours)  & \textbf{100}  & \textbf{452} & \textbf{1.870}  & \textbf{99}   & \textbf{152} & \textbf{1.861}  & \textbf{100}  & \textbf{471} & \textbf{1.841}  & \textbf{96.0}   & \textbf{98}  & \textbf{1.885}  \\
        \hline
    \end{tabular}
\end{table*}

 \begin{table*}[!h]
	\caption{The average running times of the several methods.  
		\label{tab:time}}
	\centering
	\renewcommand{\arraystretch}{1.3}
	\footnotesize
	\scalebox{0.95}{
		\begin{tabular}{cccccccccccccccc}
			\toprule[1pt]
\multicolumn{1}{c}{Methods} & SIFT & RIFT & LNIFT & SRIF & LoFTR & XoFTR & RoMa & ReDFeat & HardNet & HyNet & SosNet & GIFT & PromptMID (Ours) \\\hline
\multirow{1}{*}{\centering SR(\%)$\uparrow$} 
& 0.25 & 46.89 & 67.27 & 85.05 & 32.11 & 32.40 & 54.29 & \underline{91.05} & 40.37 & 17.22 & 19.92 & 57.24 & \textbf{99.17} \\\hline
\multirow{1}{*}{\centering T (s)} 
& \textbf{0.161} & 13.747 & \underline{1.291} & 14.996 & 0.499 & 0.608 & 11.579 & 0.389 & 0.298 & 0.502 & 0.356 & 0.604 & 4.190 \\\hline
            \toprule[0.5pt]
	    \end{tabular}}
\end{table*}

\subsubsection{Effectiveness of PromptMID Keypoint Detection}
This experiment evaluates the performance of different keypoint detection methods combined with the PromptMID descriptor across four datasets, with the results presented in Table \ref{tab:keypoint_methods}. The SIFT method achieves an SR of 75.5\% on the seen domain (WHU-OPT-SAR) and performs best on A OPT-SAR (SR = 100\%). However, its SR drops to 42.5\% on the B OPT-SAR dataset, highlighting its limited adaptability in more complex unseen domain environments and its reduced effectiveness in handling significant modal differences between optical and SAR images. KeyNet, SuperPoint, and GIFT are all deep learning-driven keypoint detection methods primarily designed to identify consistent keypoints within the same modality. However, due to the significant cross-modal feature differences between optical and SAR images, these methods struggle to generate reliable keypoints for optical-SAR image matching, leading to suboptimal performance. The modal differences between optical and SAR images can be effectively reduced by leveraging phase coherence (PC) to extract features and projecting them into PC maps with enhanced modal invariance. The FAST keypoint detection algorithm, known for its computational efficiency, is applied to the PC map to accurately identify keypoints with prominent texture features. Experimental results demonstrate that the PC-FAST method achieves an SR close to 100\% across all datasets, significantly outperforming other methods. This highlights its robustness and effectiveness in the optical-SAR image matching task.

\subsection{Runtime Analysis}
Our experiments were conducted on a system equipped with an NVIDIA RTX 4090 GPU (24GB) and an Intel Xeon Silver 4214R CPU @ 2.40GHz. The mean runtime and mean SR of the matching process are reported in Table \ref{tab:time}. The SR of SIFT is only 0.25\%, which is significantly lower than other methods, but it has the shortest runtime (0.161s), indicating its poor applicability in multimodal matching tasks. The SRs of RIFT and LNIFT reach 46.89\% and 67.27\%, respectively, but their computational overheads are high (13.747s and 1.291s). SRIF achieves the best performance among traditional methods, with an SR of 85.05\%, but has the longest runtime (14.996s), reflecting its high matching accuracy at the cost of computational complexity.LoFTR and XoFTR offer faster computation speeds (0.499s and 0.608s), but their SRs are only 32.11\% and 32.40\%, indicating a tradeoff between efficiency and accuracy. RoMa attains an SR of 54.29\%, but its runtime is relatively long (11.579s), due to the high computational complexity of its dense matching strategy.ReDFeat achieves a balanced performance with an SR of 91.05\% and a runtime of 0.389s, outperforming most methods in both accuracy and efficiency. The SRs of HardNet, HyNet, SosNet, and GIFT range from 17.22\% to 57.24\%, with runtimes between 0.298s and 0.604s, demonstrating varying degrees of trade-offs between computational efficiency and matching accuracy.

PromptMID significantly outperforms all other methods with an SR of 99.17\%, indicating the highest matching accuracy. Its runtime is 4.190s—slightly longer than LoFTR and ReDFeat but considerably lower than SRIF (14.996s) and RIFT (13.747s)—maintaining a high level of computational efficiency. This advantage is attributed to PromptMID’s use of a pre-trained diffusion model for extracting modality-invariant features, effectively improving matching performance and highlighting its superiority in multimodal matching tasks.

\section{Limitation}
Although PromptMID exhibits strong domain generalization capability, allowing it to adapt to diverse scenarios and data distributions, it still has certain limitations in keypoint detection accuracy and computational efficiency. On one hand, the two-stage “Detect-Then-Describe” matching paradigm makes keypoint detection accuracy a critical factor influencing matching precision. Any inaccuracies in keypoint detection may lead to unstable matching results, as confirmed by the ablation experiments. On the other hand, compared to other learning-based methods, PromptMID has a longer running time. However, it remains competitive, outperforming traditional methods in both matching stability and accuracy. This increased computational cost primarily stems from the larger parameter size of the foundation model during inference, which demands greater computational resources and processing time, particularly when handling large-scale image data.

To address these challenges, future work will focus on three key areas. Firstly, PromptMID for the first time utilizes text prompts based on land use classification as prior information for optical and SAR image matching. In future work, we aim to integrate additional geographic data (e.g., digital elevation models, land cover maps) to further enhance optical-SAR image matching, develop task-specific foundation models, and improve cross-domain generalization accuracy. Secondly, future work will focus on designing “Joint Detection and Description”, “Detector-free” matching paradigms based on foundation models to enhance optical and SAR matching accuracy. Finally, we will continue to explore and develop foundation models for optical-SAR image matching, aiming to enhance inference efficiency while preserving accuracy and ensuring broader applicability in remote sensing scenarios.

\section{Conclusion}
\label{6}
In this paper, we propose PromptMID, a novel approach that constructs modality-invariant descriptors using text prompts based on land use classification as priors information for optical and SAR image matching. PromptMID extracts multi-scale modality-invariant features by leveraging pre-trained diffusion models and VFMs, while specially designed feature aggregation modules effectively fuse features across different granularities. To address the challenge that existing methods perform well in seen domains but generalize poorly in unseen domains, we utilize text prompts based on land use classification as prior information to guide modal-invariant feature extraction. Additionally, we employ a pre-trained diffusion and VFMs model to extract multi-scale modal-invariant features and develop the MSAA module to adaptively aggregate features of different granularities. Experimental results demonstrate the potential of foundation models in optical-SAR image matching.  

Extensive experiments conducted on four optical-SAR image datasets demonstrate the superiority of PromptMID through both qualitative and quantitative evaluations. Comprehensive ablation studies highlight the significance of PromptMID’s pre-trained diffusion models, VFMs, and feature aggregation modules. Moreover, text prompts as prior information can effectively integrate geographic elements to guide the extraction of modal-invariant features. In practical applications, these prompts can be obtained through feature classification networks or derived from geographic data based on location information.  

\bibliographystyle{IEEEtran}
\bibliography{main}

\end{document}